# Use it or Lose it: Selective Memory and Forgetting in a Perpetual Learning Machine

Andrew J.R. Simpson [#1]

[#] *Centre for Vision, Speech and Signal Processing, University of Surrey*
*Surrey, UK*
[1] `Andrew.Simpson@Surrey.ac.uk`

*Abstract*—**In a recent article we described a new type of deep neural network– a *Perpetual Learning Machine* (PLM) – which is capable of learning 'on the fly' like a brain by existing in a state of *Perpetual Stochastic Gradient Descent* (PSGD). Here, by simulating the process of *practice*, we demonstrate both selective memory and selective *forgetting* when we introduce statistical recall biases during PSGD. Frequently recalled memories are remembered, whilst memories recalled rarely are *forgotten*. This results in a 'use it or lose it' stimulus driven memory process that is similar to human memory.**

## I. INTRODUCTION

Deep neural networks (DNN) have long aimed at replicating human intelligence but still fail to capture some important features: DNN do not learn 'on the fly', do feature emergent memory and they do not *forget*. To account for emergent memory, we recently introduced a new type of DNN – a *Perpetual Learning Machine* (PLM) – which is capable of on-the-fly learning [1]. The PLM exists in a state of *Perpetual Stochastic Gradient Descent* (PSGD) and provides a unified architecture for learning and memory represented within the weights of the model.

A remaining key difference between machine learning and human *learning* is the concept of *practice*. The maxim 'use it or lose it' invokes the role of *practice* in human learning and memory – what we do not *practice* we *forget*. Furthermore, memory is cumulative with practice – what we practice most, we remember best. Therefore, both *learning* and *forgetting* are selective processes and *practice* is the selection mechanism. This selectivity is not captured by either the standard DNN or the PLM, both of which essentially converge upon a uniform state of learning across whatever set of memorable data or classes. To restate, the concept of *practice* combines both learning and statistics – what we learn frequently, we *memorise*. Thus, since learning results in memory in the PLM, we may hypothesise that a combination of frequentist statistics and *learning* will result in selective *learning* and selective *forgetting* in the PLM.

In this article, we demonstrate that both learning and memory in a PLM may be biased by recall statistics during PSGD. Elements of the dataset which are statistically prioritised (commonly recalled) assimilate more rapidly and are remembered best during perpetual memory. Elements of the dataset not prioritised (rarely recalled) are *forgotten*.

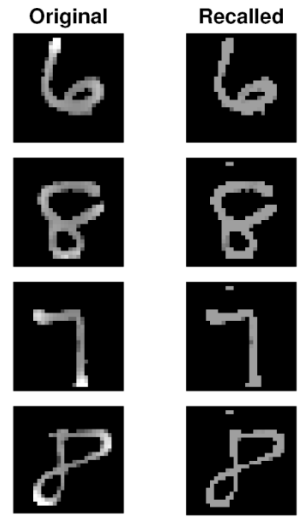

Fig. 1. **Recall of training images.** On the left are plotted MNIST digits and on the right are plotted the same digits synthesised using the *recall* DNN.

## II. METHOD

We chose the well-known MNIST hand-written digit dataset [2]. First, we unpacked the images of 28x28 pixels into vectors of length 784. Example digits are given in Fig. 1. Pixel intensities were normalized to zero mean.

Our PLM [1] involves two DNNs, one for *storage* and the other for *recall*. The *storage* DNN learns the classes of some training images. The *recall* DNN learns to synthesise the same images from the same classes. Together, the two networks hold, encoded, the training set. We then place these pair of DNNs in a self-supervised and homeostatic state of *Perpetual Stochastic Gradient Descent* (PSGD). During each step of PSGD, a random class is chosen and an image synthesised from the *recall* DNN. This randomly synthesised image is then used in combination with the random class to train both DNNs via non-batch SGD. I.e., the PSGD is driven by training data that is *synthesised* from *memory* according to random classes. In this article, we bias the statistics of this random perpetuation such that certain elements are learned more frequently than others. This allows us to measure the effect of this frequentist-statistical bias in terms of learning and memory.

*Perpetual Memory.* We required our PLM to learn to identify a collection of images. We took the first 75 of the MNIST digits and assigned each to an arbitrary class (this is arbitrary associative learning). This gave 75 unique classes, each associated with a single, specific digit. The task of the model was to recognise the images and assign to them the correct (arbitrary) classes. We split the 75 digits randomly into three groups of equal size (25). We assigned each with a different probability of being selected during PSGD.

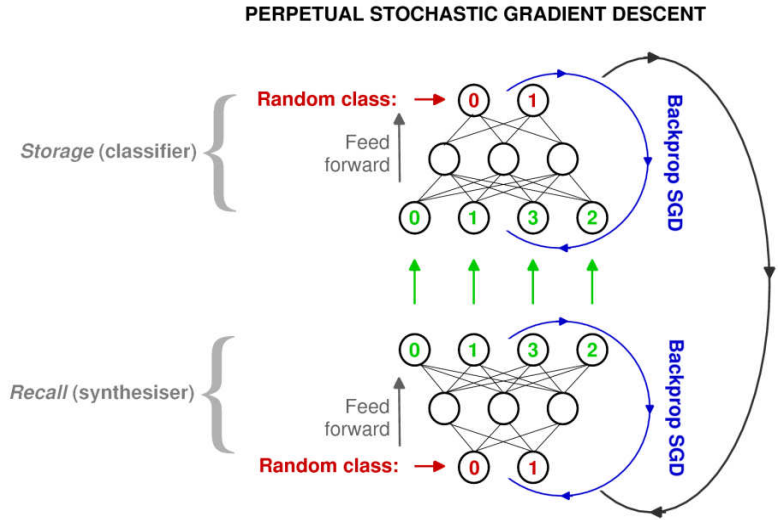

**Fig. 2. Self-supervision: PSGD schematic diagram.** For each iteration of PSGD, a random class is chosen and from this input the *recall* DNN is used to synthesise the respective training image (from memory). This recalled training image is then used with the random class to train *both* networks for a single step of backprop SGD.

*Storage and Recall.* We instantiated two DNN; the *storage* DNN was a typical classifier of size 784x100x75, with the softmax output layer corresponding to the 75-way classification problem. The *storage* DNN took images as input and produced classes as output. The *recall* DNN was of size 75x100x784, took classes as input and synthesised the training images at output. Both DNNs used biased sigmoids [3] throughout (with zero bias in the output layer).

*Selective learning.* For the *selective learning* experiment, the *storage* DNN was trained using the 75 image classes. Each step of non-batch SGD training featured only a single class (i.e., SGD training was not batch averaged). The class was randomly chosen according to the statistical biases of the groups; group 1 was chosen with 80% probability, group 2 was chosen with 15% probability and group 3 was chosen with 5% probability. Training was performed (regularised) using parallel (100x) dither w/ dropout [as in 1,4,5].

*Selective forgetting.* In the *selective forgetting* experiment, both *storage* and *recall* DNNs were independently trained (from random starting weights) on the entire 75 image classes without any statistical biases (i.e., non-batch SGD) for 100 full-sweep iterations. Classification error converged at 0.04% for the *storage* DNN, and at 0.04% for the *storage* DNN fed with the output of the *recall* DNN charged with synthesising the images of the respective test classes. Hence, the recall was suitably robust and was more or less visually indistinguishable from the original training images. Fig. 1 plots some example digits recalled (synthesised) using the *recall* DNN.

*Perpetual Stochastic Gradient Descent.* Once the *storage* and *recall* DNNs were trained, the training images were discarded and the pair of models were subjected to PSGD (Fig. 2). The random selection of classes during PSGD was subject to the statistical biases of the three groups; group 1 was selected with 99% probability, group 2 was selected with 1% probability and group 3 was not selected at all (0% probability). Next, using this selectively biased random class, a respective image was synthesised using the *recall* DNN. This synthetic image was then combined with the random class and used together to train both DNNs in parallel (via non-batch SGD [1,5]). I.e., given the random seed, the *recall* DNN synthesised – from *memory* – the relevant training image and used it for *self-supervision*. This step of non-batch SGD also employed parallel dither w/dropout (100x). As in [1,4,5], all dither was random noise of zero mean and unit scale and dropout [6,5] was 50%.

In both experiments (*selective learning* and *selective forgetting*), classification accuracy of the *storage* DNN was tested at each iteration of PSGD. Each subgroup (of 25 image classes) was tested separately. This gave three dynamic measures of memory and recall that could be plotted as a function of time (iterations).

III. RESULTS

Fig. 3a plots the recall accuracy (classification error rate) of the *storage* DNN, trained from scratch, as a function of PSGD iterations for the various groups. The group selected with 80% probability was learned fastest and the groups at 15% and 5% were learned proportionally more slowly. Hence, it is possible to introduce a selective bias to the learning phase such that commonly occuring elements are learned more quickly.

Fig. 3b plots the recall accuracy of the *storage* DNN, as a function of PSGD iterations, from the starting point of being fully trained using SGD and without any statistical biases. I.e., here, we start from the position of having memorised the whole dataset. The 99% recall group shows solid homeostasis throughout and does not deviate from zero error. Both the 1% and the 0% groups begin to increase in error (i.e., be forgotten) immediately but the 1% group recovers and reaches homeostasis at a relatively high level of error, while the 0% group continues to be increasingly forgotten without any signs of levelling off. Thus, during PSGD, the memory of the most probable image classes was maintained perfectly, whilst the less probable image classes were forgotten to a degree that was determined by their frequency of recall.

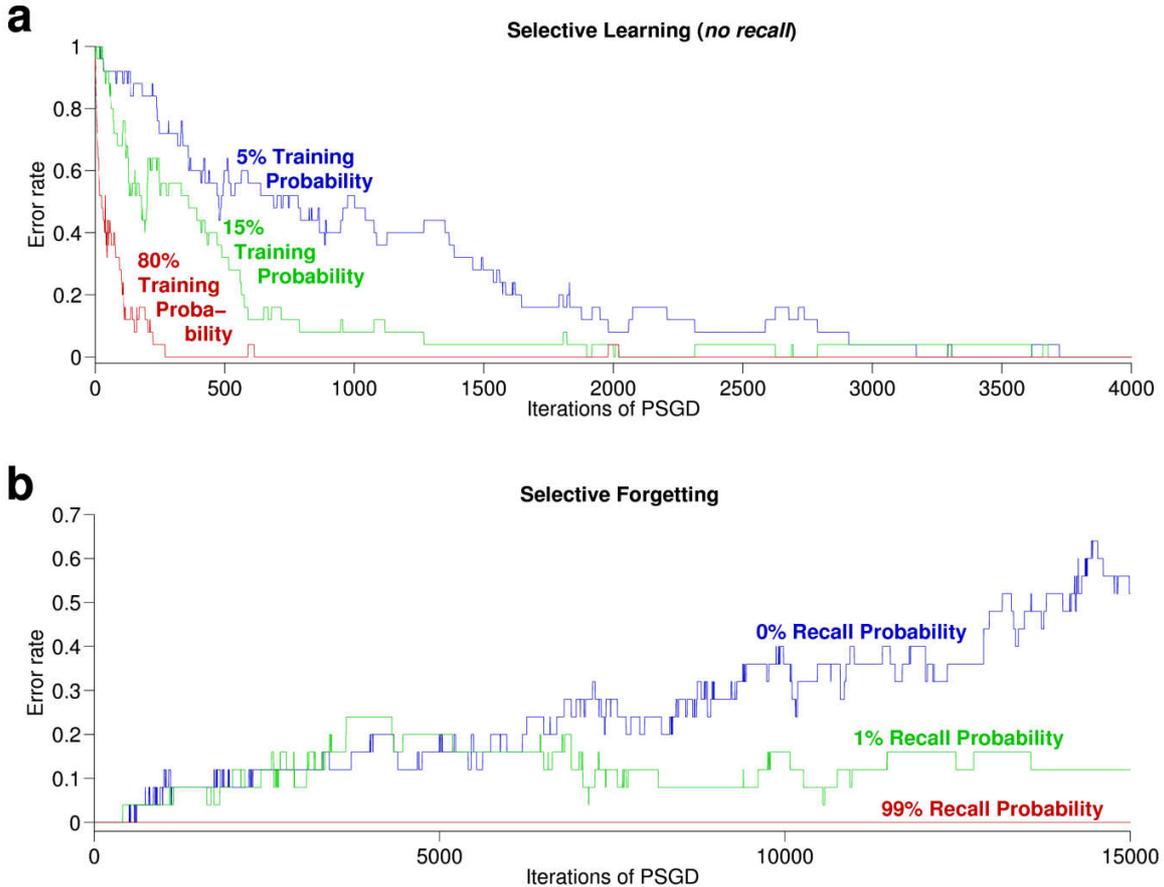

**Fig. 3. Use it or lose it: Selective learning and selective forgetting via PSGD.** The dataset (75) was split randomly into 3 groups and to each was assigned an a-priori probability of presentation. The probabilities were different for the *learning* (upper) and *forgetting* (lower) experiments. This figure plots the various *storage* DNN error functions of PSGD iterations. **a** plots error during selective learning (from scratch), illustrating the effect of the selective biases on learning for the respective subgroups. NB: *This plot begins with an untrained (storage) model*. **b** plots error during selectively-biased PSGD (i.e., PSGD from memory via recall using the *recall* DNN, see Fig. 2), illustrating the effect of 'forgetting' for the less frequently recalled groups. NB: *This plot begins with a fully trained pair of models*.

IV. DISCUSSION AND CONCLUSION

By introducing selective statistical biases into the paths of PSGD, we have demonstrated both *selective memory* and *selective forgetting* in a Perpetual Learning Machine. We have equated *learning* with stochastic gradient descent, we have equated frequentist statistics with *practice*, and we have equated the emergent and selective result on perpetual *memory* with *forgetting*. Hence, it seems possible that a similar principle may be responsible for forgetting within the brain.

ACKNOWLEDGMENT

AJRS did this work on the weekends and was supported by his wife and children.

REFERENCES

[1] Simpson AJR (2015) "On-the-Fly Learning in a Perpetual Learning Machine", arxiv.org abs/1509.00913.
[2] LeCun Y, Bottou L, Bengio Y, Haffner P (1998) "Gradient-based learning applied to document recognition", Proc. IEEE 86: 2278–2324.
[3] Simpson AJR (2015) "Abstract Learning via Demodulation in a Deep Neural Network", arxiv.org abs/1502.04042.

[4] Simpson AJR (2015) "Dither is Better than Dropout for Regularising Deep Neural Networks", arxiv.org abs/1508.04826.
[5] Simpson AJR (2015) "Parallel Dither and Dropout for Regularising Deep Neural Networks", arxiv.org abs/1508.07130.
[6] Hinton GE, Srivastava N, Krizhevsky A, Sutskever I, Salakhutdinov R (2012) "Improving neural networks by preventing co-adaptation of feature detectors", The Computing Research Repository (CoRR), abs/1207.0580.